\documentclass[11pt,a4paper]{article}
\usepackage[hyperref]{emnlp2018}
\usepackage{times}
\usepackage{latexsym}
\usepackage[final]{graphicx}
\usepackage{url}

\aclfinalcopy 

\title{Humor Detection: A Transformer Gets the Last Laugh}

\author{Orion Weller \\
  Computer Science Department \\
  Brigham Young University \\
  {\tt orionw@byu.edu} \\\And
  Kevin Seppi \\
  Computer Science Department \\
  Brigham Young University \\
  {\tt kseppi@byu.edu} }

\date{3.23.2019}
\hypersetup{final}
\begin{document}
\maketitle
 \begin{abstract}
  Much previous work has been done in attempting to identify humor in text.  In this paper we extend that capability by proposing a new task: assessing whether or not a joke is humorous.  We present a novel way of approaching this problem by building a model that learns to identify humorous jokes based on ratings gleaned from Reddit pages, consisting of almost 16,000 labeled instances.  Using these ratings to determine the level of humor, we then employ a Transformer architecture for its advantages in learning from sentence context.  We demonstrate the effectiveness of this approach and show results that are comparable to human performance. We further demonstrate our model's increased capabilities on humor identification problems, such as the previously created datasets for short jokes and puns. These experiments show that this method outperforms all previous work done on these tasks, with an F-measure of 93.1\% for the Puns dataset and 98.6\% on the Short Jokes dataset.
\end{abstract}


\section{Introduction}
Recent advances in natural language processing and neural network architecture have allowed for widespread application of these methods in Text Summarization \citep{Liu}, Natural Language Generation \citep{Bahuleyan}, and Text Classification \citep{Yang2016}.  Such advances have enabled scientists to study common language practices.  One such area, humor, has garnered focus in classification \citep{ZhangHumor, Chen}, generation \citep{he2019pun, valitutti2013let}, and in social media \citep{Raz}.  

The next question then is, what makes a joke humorous?  Although humor is a universal construct, there is a wide variety between what each individual may find humorous.  We attempt to focus on a subset of the population where we can quantitatively measure reactions: the popular Reddit r/Jokes thread. This forum is highly popular - with tens of thousands of jokes being posted monthly and over 16 million members.  Although larger joke datasets exist, the r/Jokes thread is unparalleled in the amount of rated jokes it contains.  To the best of our knowledge there is no comparable source of rated jokes in any other language.  These Reddit posts consist of the body of the joke, the punchline, and the number of reactions or \textit{upvotes}.  Although this type of humor may only be most enjoyable to a subset of the population, it is an effective way to measure responses to jokes in a large group setting.\footnote{See the thread (of varied and not safe for work content) \href{https://www.reddit.com/r/Jokes/}{at this link}. We do not endorse these jokes.}

What enables us to perform such an analysis are the recent improvements in neural network architecture for  natural language processing. These breakthroughs started with the Convolutional Neural Network \citep{LeCun} and have recently included the inception \citep{Bahdanau} and progress of the Attention mechanism \citep{Luong, Xu}, and the Transformer architecture \citep{Vaswani}. 

\section{Related Work}

In the related work of joke identification, we find a myriad of methods employed over the years: statistical and N-gram analysis \citep{Taylor}, Regression Trees \citep{Purandare}, Word2Vec combined with K-NN Human Centric Features \citep{Yang}, and Convolutional Neural Networks \citep{Chen}. 

\begin{table*}[htbp]
\centering
\begin{tabular}{p{8cm}|p{3cm}|p{3cm}}
  Body & Punchline & Score \\
  \hline
  Man, I was so tired last night; I had a dream I was a muffler... 
 & and I woke up exhausted & 276 \\
  I told my teenage niece to go get me a newspaper... She laughed at me, and said, "Oh uncle you're so old. Just use my phone." & So I slammed her phone against the wall to kill a spider. & 28315 \\
\end{tabular}
\caption{Example format of the Reddit Jokes dataset
  }
\end{table*}

This previous research has gone into many settings where humor takes place. \citet{Chen} studied audience laughter compared to textual transcripts in order to identify jokes in conversation, while much work has also gone into using and creating datasets like the Pun of the Day \citep{Yang}, 16000 One-liners \citep{Mihalcea}, and even Ted Talks \citep{Chen}.  

\section{Data}
We gathered jokes from a variety of sources, each covering a different type of humor. These datasets include jokes of multiple sentences (the Short Jokes dataset), jokes with only one sentence (the Puns dataset), and more mixed jokes (the Reddit dataset).  We have made our code and datasets open source for others to use. \footnote{Our code and datasets are publicly available at \href{https://www.github.com/orionw/RedditHumorDetection}{this link.}}

\subsection{Reddit}

Our Reddit data was gathered using Reddit's public API, collecting the most recent jokes. Every time the scraper ran, it also updated the upvote score of the previously gathered jokes. This data collection occurred every hour through the months of March and April 2019.  Since the data was already split into body and punchline sections from Reddit, we created separate datasets containing  the body of the joke exclusively and the punchline of the joke exclusively. Additionally, we created a dataset that combined the body and punchline together.


Some sample jokes are shown in Table 1, above.  The distribution of joke scores varies wildly, ranging from 0 to 136,354 upvotes.  We found that there is a major jump between the 0-200 upvote range and the 200 range and onwards, with only 6\% of jokes scoring between 200-20,000.  We used this natural divide as the cutoff to decide what qualified as a funny joke, giving us 13884 not-funny jokes and 2025 funny jokes.  

\subsection{Short Jokes}
The Short Jokes dataset, found on Kaggle, contains 231,657 short jokes scraped from various joke websites with lengths ranging from 10 to 200 characters. The previous work by \citet{Chen} combined this dataset with the WMT162 English news crawl.  Although their exact combined dataset is not publicly available, we used the same method and news crawl source to create a similar dataset. We built this new Short Jokes dataset by extracting sentences from the WMT162 news crawl that had the same distribution of words and characters as the jokes in the Short Jokes dataset on Kaggle\footnote{The Short Jokes dataset from Kaggle is available \href{https://www.kaggle.com/abhinavmoudgil95/short-jokes}{here}.}. This was in order to match the two halves (jokes and non-jokes) as closely as possible. 

\subsection{Pun of the Day}
This dataset was scraped by \citet{Yang} and contains 16001 puns and 16002 not-punny sentences. We gratefully acknowledge their help in putting together and giving us use of this dataset. These puns were constructed from the Pun of the Day website while the negative samples were gathered from news websites. 

\section{Methods}
In this section we will discuss the methods and model used in our experiments. 

\subsection{Our Model}
We have chosen to use the pre-trained BERT \citep{Devlin} as the base of our model.  BERT is a multi-layer bidirectional Transformer encoder and was initially trained on a 3.3 billion word corpus.  The model can be fined-tuned with another additional output layer for a multitude of other tasks.  We chose to use this Transformer based model as our initial platform because of its success at recognizing and attending to the most important words in both sentence and paragraph structures.

In Figure 1, originally designed by \citet{Vaswani}, we see the architecture of a Transformer model: the initial input goes up through an encoder, which has two parts: a multi-headed self attention layer, followed by a feed-forward network.  It then outputs the information into the decoder, which includes the previously mentioned layers, plus an additional masked attention step. Afterwords, it is transformed through a softmax into the output.  This model's success is in large part due to the Transformer's self-attention layers.

We chose a learning rate of 2e-05 and a max sequence length of 128. We trained the model for a maximum of 7 epochs, creating checkpoints along the way.  

\begin{figure}
\centering
\includegraphics[width=240pt, height=240pt]{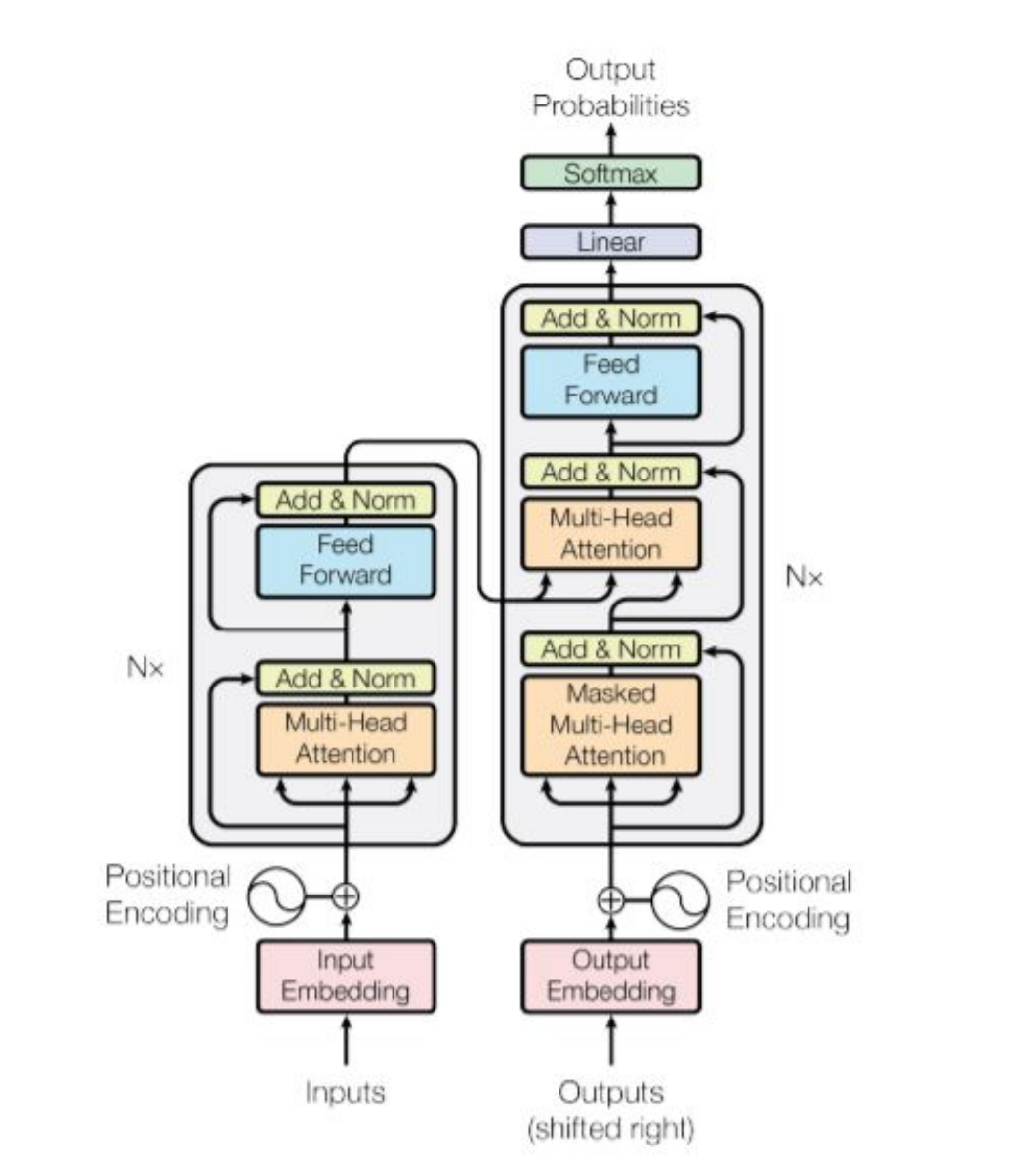}
\caption{Transformer Model Architecture}
\end{figure}

\subsection{Training}
Since our data was unbalanced we decided to upsample the humorous jokes in training.  We split the dataset into a 75/25 percent split, stratifying with the labels.  We then upsampled the minority class in the training set until it reached an even 50 percent.  This helped our model learn in a more balanced way despite the uneven amount of non-humorous jokes.  Our validation and test sets were composed of the remaining 25\%, downsampling the data into a 50/50 class split so that the accuracy metric could be balanced and easily understood.

To show how our model compares to the previous work done, we also test on the Short Joke and Pun datasets mentioned in the Data section.  For these datasets we will use the metrics (Accuracy, Precision, Recall, and F1 Score) designated in \citet{Chen} as a comparison. We use the same model format as previously mentioned, trained on the Reddit dataset.  We then immediately apply the model to predict on the Short Joke and Puns dataset, without further fine-tuning, in order to compare the model.  However, because both the Puns and Short Joke datasets have large and balanced labels, we do so without the upsampling and downsampling steps used for the Reddit dataset.

\begin{table}
\begin{tabular}{cccc}
  Method & Body & Punchline & Full \\
  \hline \\
  CNN & 0.651 & 0.684 & 0.688  \\
  Transformer &  \textbf{0.661} &  \textbf{0.692} & \textbf{0.724} \\
  Human (General) & 0.493 & 0.592 & 0.663 \\
\end{tabular}
\caption{Results of Accuracy on Reddit Jokes dataset}
\end{table}

\begin{table*}[h]
\centering
\begin{tabular}{p{4cm}cccc}
Previous Work: & Accuracy & Precision & Recall & F1 \\
\hline
Word2Vec+HCF& 0.797 & 0.776 & 0.836 & 0.705 \\
CNN & 0.867 & 0.880 & 0.859 & 0.869 \\
CNN+F & 0.892 & 0.886 & 0.907  & 0.896 \\
CNN+HN  & 0.892 & 0.889 & 0.903 & 0.896 \\
CNN+F+HN & 0.894 & 0.866 & \textbf{0.940} & 0.901 \\ 
& & & & \\
 Our Methods: & Accuracy & Precision & Recall & F1 \\ 
 \hline
Transformer & \textbf{0.930} & \textbf{0.930} & 0.931 & \textbf{0.931} \\
\end{tabular}
\caption{Comparison of Methods on Pun of the Day Dataset. $HCF$ represents Human Centric
Features, $F$ for increasing the number of filters, and $HN$ for the use of highway layers in the model. See \citep{Chen, Yang} for more details regarding these acronyms.}
\end{table*}

\section{Experiments}
In this section we will introduce the baselines and models used in our experiments. 

\subsection{Baselines}
In order to have fair baselines, we used the following two models: a CNN with Highway Layers as described by \citet{Chen} and developed by \citet{Srivastava}, and human performance from a study on Amazon's Mechanical Turk.  We wanted to have the general population rate these same jokes, thus showing the difference between a general audience and a specific subset of the population, in particular, Reddit r/Jokes users.  Since the Reddit users obviously found these jokes humorous, this experiment would show whether or not a more general population agreed with those labels. 

We had 199 unique participants rate an average of 30 jokes each with the prompt \textit{"do you find this joke humorous?"} If the participant was evaluating a sample from a body or punchline only dataset we prefaced our question with a sentence explaining that context, for example: \textit{"Below is the punchline of a joke. Based on this punchline, do you think you would find this joke humorous?"} Taking these labels, we used the most frequently chosen tag from a majority vote to calculate the percentages found in the \textit{Human} section of Table 2.


\subsection{Results}
In Table 2, we see the results of our experiment with the Reddit dataset.  We ran our models on the body of the joke exclusively, the punchline exclusively, and both parts together (labeled \textit{full} in our table). On the full dataset we found that the Transformer achieved an accuracy of 72.4 percent on the hold out test set, while the CNN was in the high 60's.  We also note that the general human classification found 66.3\% of the jokes to be humorous. 

In order to understand what may be happening in the model, we used the body and punchline only datasets to see what part of the joke was most important for humor.  We found that all of the models, including humans, relied more on the punchline of the joke in their predictions (Table 2).  Thus, it seems that although both parts of the joke are needed for it to be humorous, the punchline carries higher weight than the body. We hypothesize that this is due to the variations found in the different joke bodies: some take paragraphs to set up the joke, while others are less than a sentence.

Our experiment with the Short Jokes dataset found the Transformer model's accuracy and F1 score to be 0.986.  This was a jump of 8 percent from the most recent work done with CNNs (Table 4).

The results on the Pun of the Day dataset are shown in Table 3 above.  It shows an accuracy of 93 percent, close to 4 percent greater accuracy than the best CNN model proposed. Although the CNN model used a variety of techniques to extract the best features from the dataset, we see that the self-attention layers found even greater success in pulling out the crucial features.

\begin{table}[htb]
\small
\begin{tabular}{ccccc}
  Method & Accuracy & Precision & Recall & F1 \\
  \hline \\
  CNN+F+HN & 0.906 & 0.902 & 0.946 & 0.924 \\
  Transformer & \textbf{0.986} & 0.986 & 0.986 & 0.986 \\
\end{tabular}
\caption{Results on Short Jokes Identification }
\end{table}

\section{Discussion}

Considering that a joke's humor value is subjective, the results on the Reddit dataset are surprising.  The model has used the context of the words to determine, with high probability, what an average Reddit r/Jokes viewer will find humorous.  When we look at the general population's opinion as well, we find a stark difference between their preferences and those of the Reddit users (Table 2).  We would hypothesize that our model is learning the specific type of humor enjoyed by those who use the Reddit r/Jokes forum.  This would suggest that humor can be learned for a specific subset of the population.

The model's high accuracy and F1 scores on the Short Jokes and Pun of the Day dataset show the effectiveness of the model for transfer learning. This result is not terribly surprising. If the model can figure out which jokes are funny, it seems to be an easier task to tell when something isn't a joke at all.

Although these results have high potential, defining the absolute truth value for a joke's humor is a challenging, if not impossible task.  However, these results indicate that, at least for a subset of the population, we can find and identify jokes that will be most humorous to them.

\section{Conclusion}
In this paper, we showed a method to define the measure of a joke's humor. We explored the idea of using machine learning tools, specifically a Transformer neural network architecture, to discern what jokes are funny and what jokes are not.  This proposed model does not require any human interaction to determine, aside from the text of the joke itself, which jokes are humorous.  This architecture can predict the level of humor for a specific audience to a higher degree than a general audience consensus.  We also showed that this model has increased capability in joke identification as a result, with higher accuracy and F1 scores than previous work on this topic.




\bibliography{emnlp2018.bib}
\bibliographystyle{acl_natbib_nourl}

\appendix

\end{document}